%% file: main.tex
\documentclass[graybox]{svmult}

\usepackage{type1cm}        
\usepackage{makeidx}         
\usepackage{graphicx}        
\usepackage{multicol}        
\usepackage[bottom]{footmisc}

\usepackage{newtxtext}       %
\usepackage[varvw]{newtxmath}       

\usepackage{hyperref}
\usepackage{bm} 

\makeindex             

\input{macros}

\begin{document}

\title*{If generative AI is the answer, what is the question?}
\author{Ambuj Tewari\orcidID{0000-0001-6969-7844}}
\institute{Ambuj Tewari \at Department of Statistics, University of Michigan, 1085 S University, Ann Arbor MI 48109, USA, \email{tewaria@umich.edu}
}
%
%
\maketitle

\abstract{\input{abstract}}

\section{Overview of Generative AI}

The English word \textit{generate} traces its lineage to the Proto-Indo-European (PIE) root \textit{\'{g}enh$_1$-}, meaning ``to beget'' or ``to give birth.'' From this ancient root, we also get words like \textit{genesis}, \textit{gene}, and \textit{genre}, all of which evoke the emergence of something new from a source. The conceptual appeal of this PIE root is cross-cultural: in Sanskrit, the word \textit{ambuj}, the author's given name, combines \textit{ambu} (water) with \textit{ja} (born of), yielding ``born of water", or lotus. Generation is how we alter, enrich, and shape the world, whether in cultural, biological, artistic, or computational contexts. Generative AI reflects this long-standing legacy by giving birth to new text, images, audio, video, computer code, molecules, and even theorems. This helps explain the widespread fascination with generative AI as a technology.

But if generative AI is the answer, what is the question?\footnote{This chapter's title is inspired by Shoham, Powers, and Grenager’s 2007 paper \cite{shoham2007if} written during a formative period in the development of multi-agent learning. Generative AI today finds itself in a similarly formative stage, not in terms of capability or adoption, which are already far-reaching, but in terms of foundational understanding and theoretical frameworks.} This chapter explores how generative AI models work. We then examine the mathematical foundations of generation as a machine learning task. Finally, we consider how generation relates to other core problems such as prediction, compression, and decision-making.

\subsection{Definitions}\label{sec:defs}

It would be unreasonable to expect full consensus on the definition of an emerging field like generative AI. However, it is helpful to examine definitions from a range of sources, both public and academic, to identify common themes. For example, Wikipedia\footnote{\url{https://en.wikipedia.org/wiki/Generative\_artificial\_intelligence}} defines generative AI as:
\begin{quotation}
    ... a subfield of artificial intelligence that uses generative models to produce \emph{text, images, videos, or other forms of data}. These models learn the underlying patterns and structures of their \emph{training data} and use them to produce \emph{new data}  based on the input, which often comes in the form of natural language \emph{prompts}. (\emph{emphasis added})
\end{quotation}
An MIT news article\footnote{\url{https://news.mit.edu/2023/explained-generative-ai-1109}} explaining generative AI defines it as:
\begin{quotation}
    Generative AI can be thought of as a machine-learning model that is trained to create \emph{new data}, rather than making a prediction about a specific dataset. A generative AI system is one that learns to generate more \emph{objects that look like the data it was trained on}. (\emph{emphasis added})
\end{quotation}
Although both of these sources were written for a general audience, we can already see several recurring elements: reliance on training data, the ability to generate a variety of data types, expectation that the generated objects will resemble the training data while being novel, and the potential for user control via prompts.

We now turn to definitions found in leading textbooks written for a technical machine learning audience.  In the second volume of his authoritative textbook \emph{Probabilistic Machine Learning}~\cite{pml2Book}, Kevin Murphy devotes an entire section to generative models. He first defines a generative model as a joint probability distribution $p(\bx)$ over objects $\bx$ in some abstract space $\cX$, and then adds:
\begin{quotation}
    One of the main goals of generative models is to \emph{generate (create) new data samples}. This is sometimes called \textbf{generative AI}. For example, if we fit a model $p(\bx)$ to images of faces, we can sample new faces from it ... Similar methods can be used to create samples of \emph{text, audio, etc.} ...
    \emph{To control what is generated}, it is useful to use a conditional generative model of the form $p(\bx|\bc)$. (\emph{emphasis added})
\end{quotation}
Our second example comes from Prince’s \emph{Understanding Deep Learning}~\cite{prince2023understanding}. Although he does not directly define generative AI, he defines generative models as follows:
\begin{quotation}
Generative models can
\emph{synthesize (generate) new examples} with
\emph{similar statistics to the training data}. A
\emph{subset of these are probabilistic} and define a distribution over the data. We
draw samples from this distribution to
generate new examples. (\emph{emphasis added})
\end{quotation}
In the textbook definitions, we see many of the same themes that we saw before in popular accounts: the importance of training data, the production of novel but statistically similar outputs, the range of applicable data modalities, and the possibility of user control (often via conditioning, when using probabilistic models). Notably, Prince explicitly remarks that not all generative models are probabilistic. This underscores that the use of probabilistic models in generative AI is a modeling choice rather than an intrinsic feature of generative AI as a machine learning task. This distinction will become especially important when we examine a recent game-theoretic model of generation proposed by Kleinberg and Mullainathan~\cite{kleinberg2024language}.

Taken together, these perspectives suggest an overarching definition: generative AI is the task of producing novel outputs that respect regularities learned from data while allowing user control through prompts or conditioning. The recurring elements across definitions include:
\begin{itemize}
\item Access to training data is essential
\item Generated data can span multiple modalities, including text, images, audio, and video
\item The generated content resembles the training data
\item Novelty is expected: the outputs should not duplicate the training set
\item Potential user control over generation (e.g., via prompts or conditioning)
\end{itemize}

Any foundational theory of generation must account for these commonalities. Before discussing such foundations, we next provide an overview of the data types handled by current and emerging generative AI systems, followed by a review of the most widely used generative modeling approaches.

\subsection{Types of Data Generated}

After creating an initial wave of excitement in text and image generation, generative AI has rapidly expanded to other modalities. Some of the most commonly studied types of data include:

\begin{itemize}
    \item \textbf{Text:} Large language models such as \emph{GPT}~\cite{brown2020language} and \emph{LLaMA}~\cite{touvron2023llama} generate fluent natural language across a wide range of tasks.
    
    \item \textbf{Images:} Models like \emph{DALL-E}~\cite{ramesh2021zero} and \emph{Stable Diffusion}~\cite{rombach2022high} produce high-resolution, photorealistic images conditioned on text prompts or other structured inputs.
    
    \item \textbf{Audio and music:} Systems such as \emph{Jukebox}~\cite{dhariwal2020jukebox} and \emph{AudioLM}~\cite{borsos2023audiolm} generate music and speech.
    
    \item \textbf{Video:} Models like \emph{Sora} (OpenAI)~\cite{sora2024openai} and \emph{Veo} (Google DeepMind)~\cite{veo2024deepmind} generate high-quality video clips from text prompts, demonstrating early capabilities in modeling dynamics, scene composition, and camera motion.
    
    \item \textbf{Code:} Language models trained on code repositories, such as \emph{Codex}~\cite{chen2021evaluating} and \emph{AlphaCode}~\cite{li2022competition}, generate syntactically valid and functional programs from natural language descriptions.
    
    \item \textbf{Molecules and proteins:} \emph{Junction Tree VAE}~\cite{jin2018junction} generates valid molecular graphs, \emph{ProGen}~\cite{madani2020progen} generates protein sequences, and \emph{RFdiffusion}~\cite{watson2023novo} generates 3D protein structures.
\end{itemize}

In addition to well-established modalities, generative AI is rapidly expanding into a range of emerging domains. Models are now being developed to generate 3D shapes, physics simulations, interactive environments, and mathematical structures, supporting applications in graphics, scientific modeling, robotics, and mathematics. In domains such as healthcare and finance, generative models are exploring structured outputs such as synthetic medical records and financial time series. Finally, there is growing interest in generating structured knowledge representations, such as knowledge graphs and causal graphs, to support downstream tasks in reasoning, inference, and planning. Together, these emerging modalities signal a shift beyond perception and language, as generative AI begins to tackle increasingly interactive, structured, and abstract domains.

\subsection{Generative AI Models}

We will now discuss five major generative paradigms that all currently use neural networks as function approximators. Hence they all are deep generative paradigms since \emph{deep learning} broadly refers to the use of multi-layer neural networks for solving machine learning tasks. These five major paradigms are: autoregressive models, variational autoencoders (VAEs), normalizing flows, generative adversarial networks (GANs), and diffusion models. We will present an overview that emphasizes the basic principles used in each major generative model family. We will also point out similarities and differences as well as strengths and limitations of different models as we go along.

\subsubsection{Autoregressive Models}

Just as the chain rule of calculus drives backpropagation in deep learning, the chain rule of probability drives the success of autoregressive models. These models rely on the observation that the probability of a sequence $\bx_{1:T}$ can be decomposed as
\[
p(\bx_{1:T}) = \prod_{t=1}^T p(\bx_t \mid \bx_{1:t-1}) .
\]
This identity is exact, but in practice the set of conditioning variables $\bx_{1:t-1}$ grows with $t$, making direct modeling impractical. One classical solution is to use Markov models, which date back to the beginnings of probability theory and, via Shannon's pioneering work, to the earliest statistical models of language. A $K$-order Markov model conditions only on the most recent $K$ elements:
\[
p(\bx_t \mid \bx_{1:t-1}) \approx p(\bx_t \mid \bx_{t-K:t-1}) ,
\]
an equation that is no longer a law of probability but a modeling assumption, and a strong one if $K$ is small. However, for large $K$, the number of parameters needed to store and estimate all conditional probabilities grows exponentially with $K$.

Incorporating domain knowledge, such as sparsity or low-rank structure, can reduce the parametrization cost in high-order Markov models. Deep learning offers another route: use a neural network $f_\theta$ to map $\bx_{t-K:t-1}$ to a distribution over the output space $\cX$. When $\cX$ is discrete, $f_\theta$ outputs a categorical distribution $f_\theta(\cdot \mid \bx_{t-K:t-1})$ giving us a neural autoregressive model:
\[
p(\bx_t \mid \bx_{t-K:t-1}) = f_\theta(\bx_t \mid \bx_{t-K:t-1}) .
\]
Depending on the application, $f_\theta$ may be fully connected, convolutional, or based on the transformer architecture.

Autoregressive models using transformers have become the dominant approach to language modeling, and virtually all large language models (LLMs) follow this design. We will not delve into the details of the attention mechanism~\cite{vaswani2017attention} here, as our focus is on generative modeling rather than specific architectures. At a high level, transformers consist of multiple layers, with self-attention layers enabling the model to capture long-range dependencies. The number of parameters in these layers is largely independent of the context length $K$ (the analogue of Markov order), although training and generation (often called ``inference'' in the ML literature, but distinct from statistical inference) typically require $O(K^2)$ computation. Variants with more efficient attention mechanisms can reduce this cost. The introduction of transformers into autoregressive modeling has allowed $K$ to reach hundreds of thousands, and in some experimental systems, even millions, an impressive engineering achievement.

Since autoregressive models define a full probability distribution, they can be trained using maximum likelihood estimation (MLE). For a single sequence, the negative log likelihood (NLL) averaged over the sequence is
\[
- \frac{1}{T} \sum_{t=1}^T \log p(\bx_t \mid \bx_{1:t-1}) .
\]
In many LLM papers, the reported metric is \emph{perplexity} (PPL), which (assuming base-2 logarithms) is two raised to the power of the average log loss:
\[
\text{PPL} = \left( \prod_{t=1}^T \frac{1}{p(\bx_t \mid \bx_{1:t-1})} \right)^{1/T} .
\]
Thus, perplexity is the geometric mean of the inverse probabilities assigned to the sequence elements $\bx_t$ by the model $p$.
However, perplexity does not necessarily correlate with human-perceived quality of generated text, and models with similar perplexities may differ substantially in sample quality.
Using a neural parameterization with Markov order $K$, the optimization problem becomes
\[
\min_{\theta} \ - \frac{1}{T} \sum_{t=1}^T \log f_\theta(\bx_t \mid \bx_{t-K:t-1}) ,
\]
which, as in most deep learning applications, is solved using some form of stochastic gradient descent. 

The normalized log loss is closely connected to information-theoretic quantities. For example, if the true data-generating process $p^\star$ is stationary and ergodic, the Shannon--McMillan--Breiman theorem gives the almost-sure limit
\[
\lim_{T\to\infty} - \frac{1}{T} \sum_{t=1}^T \log p^\star(\bx_t \mid \bx_{1:t-1})
= H^\star ,
\]
where $H^\star$ is the \emph{entropy rate}. For a stationary process (no ergodicity needed), the entropy rate is well defined as the block-entropy limit
\[
H^\star := \lim_{T \to \infty} \frac{H(\bx_{1:T})}{T},
\]
and it admits the equivalent form
\[
H^\star = \lim_{T\to\infty} H(\bx_T \mid \bx_{1:T-1}) \ .
\]
See, e.g., \cite[Sec.~4.2]{cover2006elements}.

Once we have a fitted autoregressive model $p$, neural or otherwise, it can be used for generation in a straightforward way. First, sample $\hat\bx_1$ from $p(\cdot \mid \emptyset)$ (conditioning on the empty sequence), then iteratively sample $\hat\bx_t$ from $p(\cdot \mid \hat\bx_{1:t-1})$ for $t \ge 2$. During training, conditioning is on $\bx_{1:t-1}$, the \emph{true} observations from the data source. This is known as \emph{teacher forcing}. During generation, however, conditioning is on $\hat\bx_{1:t-1}$, the model's own past predictions. This mismatch is known as \emph{exposure bias} and can cause degradation in generation quality. One mitigation is \emph{scheduled sampling}~\cite{bengio2015scheduled}, where the model is gradually exposed to its own predictions during training by replacing ground-truth tokens with predicted ones with increasing probability.

Finally, the simplest decoding strategy is \emph{random sampling}, where each token is drawn from the model's predicted distribution at each step. A \emph{decoding strategy} more generally refers to any method for producing a generated sequence $\hat\bx_{1:T}$ from a trained autoregressive model. Common strategies include:
\begin{itemize}
    \item \textbf{Greedy decoding:} Select the most likely token at each step, i.e., 
    \[
    \hat\bx_t = \arg\max_{\bx \in \mathcal{X}} p(\bx \mid \hat\bx_{1:t-1}) .
    \]
    \item \textbf{Top-$k$ sampling:} Restrict sampling to the $k$ most probable tokens at each step, then renormalize probabilities over this set before sampling.
    \item \textbf{Nucleus (top-$p$) sampling:} Restrict sampling to the smallest set of tokens whose cumulative probability exceeds $p$, then renormalize over this set before sampling.
\end{itemize}

It is useful to place autoregressive models within a broader view of how deep generative models produce samples. Variational autoencoders, generative adversarial networks, and normalizing flows all begin by drawing an input vector $\bz$ from a simple distribution such as a standard Gaussian. In VAEs, $\bz$ has the interpretation of a latent variable in a probabilistic model, whereas in GANs and flows it is simply a source of input noise. This vector is then transformed deterministically through a learned network into an output in the data space, so randomness enters only once, at the start of generation. By contrast, autoregressive models compute a conditional probability distribution over the next element given the past and introduce randomness only at the end, when these probabilities are converted into actual outputs step by step. Score-based models, including diffusion models and flow matching, follow yet another pattern: they start from a random noise sample in the data space and gradually transform it into a clean output through a sequence of (often stochastic) denoising steps. This ``where the randomness enters'' perspective provides a unifying way to compare these model families, and we will return to it in the sections that follow.

\subsubsection{Variational Autoencoders (VAEs)}

As we discussed above, VAEs~\cite{kingma2014auto} generate by first sampling a latent variable $\bz$ from a tractable distribution such as a standard multivariate Gaussian $\stdnormal$. Then a neural network $f_\theta(\bz)$ (often called the \textbf{decoder}) is used to model $p( \bx \mid \bz)$, e.g., $\bx$ given $\bz$ has the distribution $p_\theta(\bx \mid \bz) = \cN(f_\theta(\bz), \sigma^2 \eye)$. Having learned $\theta$, generation is extremely efficient: sample $\bz$, compute $f_\theta(\bz)$ and sample $\bx \sim \cN(f_\theta(\bz), \sigma^2 \eye)$. For simplicity, we treat $\sigma^2$ as fixed although in practice it might be learned from data. We mainly focus on how to learn $\theta$ from data.

We do have a well-defined likelihood
\[
p_\theta(\bx) = \int_\bz p(\bz) p_\theta(\bx \mid \bz) \, d\bz
\]
which means that, given iid draws $\bx_{1:n}$, we can estimate $\theta$ by solving
\[
\max_\theta \sum_{i=1}^n \log p_\theta(\bx_i)\ .
\]
However, the complex nonlinearities in $f_\theta(\bz)$ mean that neither the objective function nor its gradient can be written in closed form or computed easily. So we give up on maximum likelihood estimation. The key idea in VAEs is to use a variational representation of $\log p_\theta(\bx)$. The Donsker-Varadhan variational principle gives the exact identity
\[
\log \EE_{Q}{[\,\exp(f)]} = \sup_{R}\, \EE_{R}{[f]}
- \KL\!\big( R\, \| \, Q \big) \ .
\]
where the supremum is over all $R$ absolutely continuous w.r.t. $Q$. The maximizer is given by $R^\star \propto Q \exp(f)$. Using the DV principle with $f = \log p_\theta(\bx|\bz)$, $Q = p(\bz)$ and $R = q(\bz)$, we have
\begin{align*}
\log p_\theta(\bx) &= \log \EE_{p(\bz)}[ \exp(\log p_\theta(\bx|\bz)) ] \\
&= \sup_{q}\Big\{\; \EE_{q(\bz)}[\log p_\theta(\bx\mid \bz)] \;-\; \KL\!\big(q(\bz)\,\|\,p(\bz)\big) \;\Big\},
\end{align*}
where the supremum is over all $q$ absolutely continuous w.r.t.\ $p$. The maximizer is the true posterior $q^\star(\bz)=p_\theta(\bz\mid \bx)$, hence
\[
\log p_\theta(\bx)
= \EE_{p_\theta(\bz\mid \bx)}[\log p_\theta(\bx\mid \bz)] \;-\; \KL\!\big(p_\theta(\bz\mid \bx)\,\|\,p(\bz)\big).
\]

The next step is to lower bound the supremum using a more restricted class that $q$ ranges over. In principle, one could use a different parameterized network for each training data point $\bx_i$. But following the amortized variational inference literature, we restrict $q$ to a parametric variational family $q_\phi(\bz\mid \bx) = \cN(\mu, \Sigma)$ where both $\mu, \Sigma$ are computed using a second neural network $g_\phi(\bx)$ (often called the \textbf{encoder}) with its own set of parameters $\phi$. We maximize only over $\phi$:
\[
\log p_\theta(\bx)
\;\;\ge\;\; 
\max_{\phi}\Big\{\EE_{q_\phi(\bz\mid \bx)}[\log p_\theta(\bx\mid \bz)] - \KL\!\big(q_\phi(\bz\mid \bx)\,\|\,p(\bz)\big)\Big\}
.
\]
The objective inside the braces is the \emph{evidence lower bound} (ELBO),
\[
\mathcal L(\theta,\phi;\bx) =
\EE_{q_\phi(\bz\mid \bx)}[\log p_\theta(\bx\mid \bz)] - \KL\!\big(q_\phi(\bz\mid \bx)\,\|\,p(\bz)\big),
\]
and maximizing it over a \emph{smaller} family (parametric $q_\phi$ rather than all $q$) yields a lower bound on $\log p_\theta(\bx)$. Although ELBO is a lower bound, it differs from the true log-likelihood exactly by the KL divergence between the variational posterior and the true posterior. That is,
\[
\log p_\theta(\bx)
= \mathcal L(\theta,\phi;\bx) \;+\; \KL\!\big(q_\phi(\bz\mid \bx)\,\|\,p_\theta(\bz\mid \bx)\big) \ .
\]
So the bound is tight iff $q_\phi(\bz\mid \bx)=p_\theta(\bz\mid \bx)$ almost everywhere. For a dataset $\bx_{1:n}$ we maximize $\sum_i \mathcal L(\theta,\phi;\bx_i)$ using stochastic gradient ascent, as the gradients with respect to $\theta$ and $\phi$ can be computed by automatic differentiation engines.

The basic VAE has been modified in many ways. Two variants that deserve mention are:
\begin{itemize}
\item $\beta$-VAE: introduces a weight $\beta$ on the KL term to encourage disentangled latent representations.
\item Vector-quantized VAE (VQ-VAE): uses a discrete codebook for $\bz$ with nearest-neighbor quantization, enabling high-quality discrete latent representations.
\end{itemize}

VAEs maximize a tractable lower bound on the log-likelihood and produce samples by decoding from a tractable latent prior. In the next sections, we explore alternative approaches: normalizing flows, which preserve exact likelihood computation by design, and generative adversarial networks, which dispense with likelihoods altogether in favor of a game-theoretic training objective.

\subsubsection{Normalizing Flows}

Just like VAEs, normalizing flows are in the ``randomness enters first" camp where we start with a simple structured random variable $\bz$ with base distribution $p_\bz(\bz)$ (which is often standard multivariate normal) and apply neural transformations to it to create a generated sample.

Suppose we apply a transformation $f$ to $\bz$ to get $\bx$. That is $\bx = f(\bz)$. Assuming $f$ is differentiable and invertible with a differentiable inverse, we have the following change of variables formula for the density of $\bx$:
\[
p_\bx(\bx) = p_\bz( f^{-1}(\bx)) \cdot | \det Df^{-1}[\bx] | \ ,
\]
where $Df^{-1}$ is the Jacobian of the inverse mapping $f^{-1}$. This formula can help us perform exact maximum likelihood estimation provided we can ensure the invertibility and differentiability conditions on $f$ are met and we can efficiently compute all the required Jacobians and their determinants. In normalizing flows, the mapping $f = f_\theta$ is parameterized by neural networks and $\theta$ can be fitted via MLE as follows:
\[
\max_{\theta}\ \sum_{i=1}^n \log p_\theta(\bx_i)
= \max_{\theta}\ \sum_{i=1}^n \big( \log p_\bz(f_\theta^{-1}(\bx_i))
+ \log | \det D f_\theta^{-1} (\bx_i) | \big) \ .
\]
Unlike the VAE case, here the likelihood is exact. However, normalizing flows do not maximize it exactly. Instead the negative log likelihood is minimized using (stochastic) gradient descent as is the norm in deep learning.

Because deep neural networks are compositions of nonlinear mappings, we have that $f(\bz) = f_L \circ \cdots \circ f_1(\bz)$ where we have suppressed the dependence on parameters $\theta$ to reduce clutter. If each of the $L$ layers consists of differentiable and invertible maps then we have
\[
f^{-1}(\bx) = f_1^{-1} \circ \cdots \circ f_{L}^{-1}(\bx) \ .
\]
By the chain rule of calculus, we can then compute the Jacobian $D f^{-1}$ at $\bx$ as:
\[
D f^{-1}(\bx) = Df_1^{-1}(\bu_1) \, Df_2^{-1}(\bu_2) \cdots Df_L^{-1}(\bu_L)
\]
where $\bu_L = \bx$ and $\bu_{i} = f_{i+1}^{-1}(\bu_{i+1})$ for $i = L-1,\ldots,1$. Therefore, the log absolute determinant of the Jacobian becomes a sum:
\[
\log | \det D f^{-1}(\bx) |
= \sum_{j=1}^L \log | \det D f_j^{-1} (\bu_j) |
\]

The theory of normalizing flows is simple, relying only on the change of variables formula and the chain rule to permit exact likelihood calculation which is used at training time to maximize the likelihood of training data. At inference, samples can be generated from the base random variable $\bz$ (often a standard multivariate normal) by applying $f = f_L \circ \cdots \circ f_1$ to $\bz$. Moreover, given a point $\bx$ in the generated space, the exact density can be calculated using the base density and Jacobians as shown above.

However, for making this simple idea work, the layers $f_j$ of the network need to be carefully designed to meet several desired characteristics. First, together the layers need to be sufficiently expressive to transform a standard multivariate normal density into complex density functions. Second, the inverse $f_j^{-1}$ should be well defined and easy to compute. Third, we should be able to efficiently compute the Jacobian of $f_j$ or $f_j^{-1}$ easily.

One way to satisfy all these different criteria is to start with simple flows like linear flows or elementwise flows and then using them as building blocks to create more complex invertible layers, or flows, such as coupling flows, autoregressive flows, and residual flows. For more details on these flows the reader can refer to \cite[Chapter 16]{prince2023understanding} and \cite[Chapter 23]{pml2Book}. 

The strengths of normalizing flows are that they support both generation and density evaluation. However, the complex requirements on invertible layers can limit expressivity. Like autoregressive models, we can write down the exact likelihood for normalizing flows. They map a base random variable $\bz$ to generated sample $\bx$ just like VAEs although the latter usually have $\bz$ in a much lower dimensional space than $\bx$. If we want to have a smaller latent variable $\bz$, we can no longer insist on invertible layers. Our next family, GANs, sticks with generation using a latent $\bz$ but completely avoids likelihoods or even approximations to the likelihood. Instead it uses a separate network, the discriminator, to learn how to generate.

\subsubsection{Generative Adversarial Networks (GANs)}

All the generative methods we have seen so far are based directly or indirectly on the principle of maximum likelihood. Generative Adversarial Networks (GANs) represent a fundamentally different approach. 
Like VAEs and normalizing flows, they start with a latent variable $\bz$ drawn from a simple base distribution 
(such as $\stdnormal$) with density $p_\bz(\cdot)$. A generator network $G_\theta$ transforms $\bz$ into a sample $\bx = G_\theta(\bz)$. 
However, unlike VAEs or normalizing flows, we do not have any explicit density $p_\theta(\bx)$ associated with the generator. 
Instead, GAN training relies on an auxiliary network called the discriminator.

The discriminator $D_\phi$ is a probabilistic classifier trained to distinguish between real samples from the data distribution 
$p_{\text{real}}$ and the samples produced by the generator. We interpret $D_\phi(\bx)$ as the probability that the candidate $\bx$ is real. So the discriminators job is to make $D_\phi$ close to one on real data examples and close to zero on generated examples.
The generator is trained simultaneously to produce samples to make the discriminator's job hard. 
So instead of relying on simple optimization like maximum likelihood, GANs rely on solving the following two-player minimax game:
\[
\min_\theta \max_\phi \ 
\mathbb{E}_{\bx \sim p_{\text{real}}} [\log D_\phi(\bx)] \ + \ 
\mathbb{E}_{\bz \sim p_\bz} [\log (1 - D_\phi(G_\theta(\bz)))] \ .
\]

This minimax formulation is due to Goodfellow et al.~\cite{goodfellow2014generative}. We can give it a nice theoretical interpretation if maximize not over parameterized class $D_\phi$ but over all classifiers $D$.
One can show that the optimal discriminator $D^\star$ is
\[
D^\star(\bx) = \frac{p_\text{real}(\bx)}{p_\text{real}(\bx) + p_\theta(\bx)} \ .
\]
Note that this is a theoretical construct since the density $p_\theta(\cdot)$ of the generated data $G_\theta(\bz)$ cannot be calculated explicitly in any efficient way. However, if we plug this optimal unconstrained discriminator into the minimax formulation, we get
\[
\min_\theta \ 2 \cdot \text{JS}\!\left( p_{\text{real}} \ \Vert \ p_\theta \right) - \log 4 \ .
\]
Here $\text{JS}$ denotes the Jensen-Shannon divergence:
\[
\mathrm{JS}(P \,\Vert\, Q) 
\;=\; \tfrac{1}{2}\,\KL\!\left(P \,\Vert\, M\right) 
\;+\; \tfrac{1}{2}\,\KL\!\left(Q \,\Vert\, M\right),
\]
where $M = \tfrac{1}{2}(P+Q)$ is the mixture distribution. Of course if $G_\theta$ also had infinite model capacity then the minimum would be achieved at $p_\theta = p_{\text{real}}$ in which case the optimal discriminator would just be $D^\star(\bx) = \tfrac{1}{2}$.

However, in practice, neither $G_\theta$ nor $D_\phi$ has infinite capacity and optimization proceeds by alternating between updates to $\theta$ and $\phi$. The dynamics of these alternating updates can be unstable. GANs can also suffer from a problem known as \emph{mode collapse} where the generator only produces a limited variety of samples instead of sampling over the full data generating distribution. Despite these limitations, GANs were very popular for generating samples, especially images, as they produced more realistic samples with better perceptual qualities compared to VAEs and normalizing flows. Numerous variants of GANs exist including ones that replace the JS divergence with other distance measures over probability distributions such as Wasserstein distance.

To summarize, GANs break with the maximum likelihood paradigm entirely, formulating the problem as a two-player game between generator and discriminator. While groundbreaking in terms of sample quality, their limitations include the lack of explicit likelihoods and well-known training instabilities. Several variants, such as Wasserstein GAN (WGAN)~\cite{arjovsky2017wasserstein} and Least-Squares GAN (LS-GAN)~\cite{mao2017least}, modify the divergence or loss function to improve stability and mitigate mode collapse. This sets the stage for the most recent family of generative models, diffusion models, which combine likelihood-based training with remarkable sample quality and training stability.

\subsubsection{Diffusion Models}

Diffusion models mark a return to maximum likelihood training but offer a very different generative approach based on stochastic noising and denoising. Like VAEs, normalizing flows, and GANs, they map a latent with simple base distribution to data. What
distinguishes them is \emph{how} the model is trained and how likelihood is handled: instead of an
explicit change-of-variables (flows) or an adversarial game (GANs), diffusion corrupts data with noise
and learns to reverse that corruption with a stable, regression-like objective. Like flows, they use a latent with the same dimensionality as data. Like VAEs, they maximize not the likelihood but a tractable variational lower bound.

Let us first define the \emph{forward process}.\footnote{In contrast to normalizing flows where ``forward" refers to the forward pass in the neural network mapping latents to data, in diffusion the ``forward" process maps data to latents.} The forward process is sometimes called the \emph{encoder} in analogy with VAEs. However, unlike VAEs, the encoder in diffusion models has no trainable parameters and it does not map to a lower dimensional space.
The forward process is governed by the following Markov chain that gradually adds noise to data given a noise schedule $\beta_t \in (0,1)$:
\[
q(\bx_t \mid \bx_{t-1}) \;=\; \mathcal{N}\!\big(\sqrt{1-\beta_t}\,\bx_{t-1},\,\beta_t \mathbf{I}\big) \ .
\]
If we define
\[ \alpha_t = 1-\beta_t,\ \ \bar\alpha_t = \prod_{s=1}^t \alpha_s \ ,
\]
then $q(\bx_t \mid \bx_0)$ has the closed form
\[
q(\bx_t \mid \bx_0) \;=\; \mathcal{N}\!\big(\sqrt{\bar\alpha_t}\,\bx_0,\,(1-\bar\alpha_t)\mathbf{I}\big) \ .
\]
For a noise schedule and large enough $T$ chosen such that $\bar\alpha_T \approx 0$, both
\(q(\bx_T)\) and \( q(\bx_T \mid \bx_0) \) are close to \(\stdnormal\).

In the \emph{reverse diffusion process}, or \emph{decoder}, we try to invert the forward diffusion process.
If we know $\bx_0$, then we once again have the closed form
\[
q(\bx_{t-1} \mid \bx_t, \bx_0) \;=\; \mathcal{N}\!\big( \, \tilde{\mu}(\bx_t, \bx_0)
, \, \tilde\beta_t \mathbf{I}\big) \ ,
\]
where
\[
\tilde\mu( \bx_t, \bx_0) =
\frac{\sqrt{\bar\alpha_{t-1}} \, \beta_t}{1-\bar\alpha_t}\bx_0 + \frac{\sqrt{\bar\alpha_t}\,(1-\bar\alpha_{t-1})}{1-\bar\alpha_t}\bx_t,
\quad
\tilde\beta_t = \frac{1-\bar\alpha_{t-1}}{1-\bar\alpha_t} \beta_t \ .
\]
Of course, at generation time we do not have $\bx_0$. So we model the reverse transitions with a Gaussian family
\[
p_\theta(\bx_{t-1}\mid \bx_t) \;=\; \mathcal{N}\!\big(\boldsymbol{\mu}_\theta(\bx_t,t),\,\sigma_t^2 \mathbf{I}\big),
\]
where natural choices for $\sigma_t^2$ are $\beta_t$ or $\tilde\beta_t$. The joint distribution $p_\theta(\bx_{0:T})$ is given by $p(\bx_T) \prod_{t=1}^T p(\bx_{t-1} \mid \bx_t)$ where $p(\bx_T)$ is $\stdnormal$. Samples $\bx_0$ can be drawn by sampling $\bx_T \sim \stdnormal$ and then recursively drawing $\bx_{t-1} \sim \mathcal{N}\!\big(\boldsymbol{\mu}_\theta(\bx_t,t),\,\sigma_t^2 \mathbf{I}\big)$ for $t=T,\dots, 1$.

For fitting the model, we have the log likelihood:
\begin{align*}
\log p_\theta(\bx_0) &= \log \int_{\bx_{1:T}} p_\theta(\bx_{0:T}) d\bx_{1:T} \\
&= \log \int_{\bx_{1:T}} q(\bx_{1:T}|\bx_0)
\frac{p_\theta(\bx_{0:T})}{q(\bx_{1:T}|\bx_0)} d\bx_{1:T} \\
&= \log \EE_{q(\bx_{1:T}|\bx_0)}\left[ \frac{p_\theta(\bx_{0:T})}{q(\bx_{1:T}|\bx_0)} \right] \\
&\ge \EE_{q(\bx_{1:T}|\bx_0)}\left[ \log \frac{p_\theta(\bx_{0:T})}{q(\bx_{1:T}|\bx_0)} \right] \\
&= \EE_{q(\bx_{1:T}|\bx_0)}\left[ \log p(\bx_T) \prod_{t=1}^T \frac{p_\theta(\bx_{t-1}|\bx_{t})}{q(\bx_{t}|\bx_{t-1})} \right]
\end{align*}
where the inequality is to due to concavity of log and Jensen's inequality. As in the VAE case, we do not directly maximize the log likelihood but rather this evidence lower bound (ELBO) that we just derived. However, note that in the VAE case, the ELBO involved another trainable distribution $q_\phi$ parameterized using neural networks. Here in the diffusion case, $q$ is fixed and depends on the forward noise adding process.

Using the Markov property of the forward process and the Bayes rule, we have
\[
q(\bx_{t}|\bx_{t-1}) = q(\bx_{t}|\bx_{t-1}, \bx_0)
\propto q(\bx_{t-1} | \bx_t, \bx_0)
\]
where the proportionality constant is independent of $\theta$. Therefore, up to terms independent of $\theta$, the ELBO is
\begin{align*}
&\quad \EE_{ q(\bx_{1:T}|\bx_0)}\left[ \log  \prod_{t=1}^T \frac{p_\theta(\bx_{t-1}|\bx_{t})}{q(\bx_{t-1}|\bx_{t}, \bx_0)} \right] \\
&= \EE_{ q(\bx_{1:T}|\bx_0)}\left[ \sum_{t=2}^T \log \frac{p_\theta(\bx_{t-1}|\bx_{t})}{q(\bx_{t-1}|\bx_{t}, \bx_0)} + \log p_\theta(\bx_0|\bx_1) \right] \ .
\end{align*}
At this point we can drop the single term outside the summation since it is not likely to significantly affect the optimization. This simplification is standard in diffusion models because the $\log p_\theta(\bx_0|\bx_1)$ term contributes very little under typical noise schedules, and omitting it does not materially change the learned reverse process~\cite{ho2020denoising}. Focusing on one of the terms inside the summation and negating it gives us the loss
\[
L_{t-1}(\theta) = -\EE_{ q(\bx_{1:T}|\bx_0)}
\left[
\log \frac{p_\theta(\bx_{t-1}|\bx_{t})}{q(\bx_{t-1}|\bx_{t}, \bx_0)}
\right]
= \EE\left[ \KL\!\big(q(\bx_{t-1}\!\mid\!\bx_t,\bx_0)\,\Vert\,p_\theta(\bx_{t-1}\!\mid\!\bx_t)\big)
\right]
\]
Note that the forward Markov chain is linear–Gaussian,
\[
\bx_t \;=\; \sqrt{\bar\alpha_t}\,\bx_0 \;+\; \sqrt{1-\bar\alpha_t}\,\bepsilon,
\quad \bepsilon\sim \stdnormal,
\]
and the true posterior is Gaussian:
\[
q(\bx_{t-1}\mid \bx_t,\bx_0)=\mathcal{N}\!\big(\tilde\mu(\bx_t,\bx_0),\,\tilde\beta_t\,\eye\big),
\quad
\tilde\mu(\bx_t,\bx_0)
=\frac{1}{\sqrt{\alpha_t}}\!\left(\bx_t-\frac{\beta_t}{\sqrt{1-\bar\alpha_t}}\bepsilon\right).
\]
Thus when we model the reverse conditional as
$p_\theta(\bx_{t-1}\mid \bx_t)=\mathcal{N}\!\big(\mu_\theta(\bx_t,t),\,\sigma_t^2\eye\big)$, it is convenient to choose the \emph{noise prediction} parametrization
\[
\mu_\theta(\bx_t,t)
=\frac{1}{\sqrt{\alpha_t}}
\left(\bx_t-\frac{\beta_t}{\sqrt{1-\bar\alpha_t}}\ \bepsilon_\theta(\bx_t,t)\right).
\]
For simplicity assume that we choose $\sigma_t^2=\tilde\beta_t$, the true posterior variance,
then for Gaussians with equal covariance the per-step KL reduces to a quadratic in the
difference of means:
\[
\KL\!\big(q(\bx_{t-1}\!\mid\!\bx_t,\bx_0)\,\Vert\,p_\theta(\bx_{t-1}\!\mid\!\bx_t)\big)
=\frac{1}{2\sigma_t^2}\,\EE\!\left[\ \|\tilde\mu(\bx_t,\bx_0)-\mu_\theta(\bx_t,t)\|_2^2\ \right].
\]
Substituting the two means and using the closed form for $\bx_t$ gives a \emph{weighted}
noise-prediction loss:
\[
L_{t-1}(\theta)
=\frac{\beta_t^2}{2\,\sigma_t^2\,\alpha_t\,(1-\bar\alpha_t)}\ 
\EE\!\left[\ \|\bepsilon-\bepsilon_\theta(\bx_t,t)\|_2^2\ \right].
\]
Following Ho et al.~\cite{ho2020denoising}, we drop the weights in front of the expectation and optimize the
\emph{simple loss}
\[
\mathcal{L}_{\text{simple}}(\theta)
=\EE_{t\sim\mathrm{Unif}\{1{:}T\},\ \bx_0\sim q_0,\ \bepsilon\sim\stdnormal}
\ \big[\ \|\bepsilon-\bepsilon_\theta(\bx_t,t)\|_2^2\ \big],
\]
where $q_0(\bx_0) = p_\text{real}(\bx_0)$ is the empirical data distribution. Optimizing this simple regression-like loss yields stable training and high sample quality.

Despite their successes, diffusion models have limitations. 
The most important practical issue is \emph{sampling speed}: 
generation requires hundreds of reverse steps, 
though there has been significant progress in reducing this cost. 

It is instructive to place diffusion models in the broader landscape of deep generative paradigms. 
Autoregressive models offer exact likelihoods but slow, sequential generation. 
VAEs are fast and flexible but optimize only a variational bound and may suffer in sample quality.
Normalizing flows provide exact likelihoods with tractable Jacobians but are limited by invertibility constraints. 
GANs break with maximum likelihood entirely and achieve sharp samples 
through adversarial training, but at the cost of training instability and lack of 
explicit likelihoods. 
Diffusion models strike a compelling balance: 
they return to likelihood-based training like VAEs and flows, 
achieve sample quality rivaling or surpassing GANs, 
and are distinguished by their remarkably stable training dynamics.

\begin{table}[h]
\centering
\small
\begin{tabular}{p{2.7cm} p{3.5cm} p{1.8cm} p{3.2cm}}
\hline
\textbf{Model} & \textbf{Training Objective} & \textbf{Likelihood} & \textbf{Typical Issues} \\
\hline
Autoregressive  &
MLE via factorization &
Exact &
Sequential sampling; exposure bias \\
\hline
VAEs &
ELBO maximization &
Approximate  &
Posterior collapse; blurry samples \\
\hline
Normalizing Flows &
MLE via change of variables &
Exact &
Invertibility requirements; expressivity limits\\
\hline
GANs &
Adversarial minimax &
Implicit &
Instability; mode collapse \\
\hline
Diffusion Models &
Variational bound; denoising & 
Implicit &
Slow sampling; depends on noise schedule \\
\hline
\end{tabular}
\caption{Comparison of major generative model families.}
\label{tab:gen-model-compare}
\end{table}

\subsubsection{Explicit vs. Implicit Likelihood Models}

A useful way to organize the five model families above is by whether they define an 
\emph{explicit} likelihood. Autoregressive models and normalizing flows provide exact 
likelihoods, and VAEs optimize a variational lower bound. In contrast, GANs and diffusion 
models define \emph{implicit} distributions and do not yield tractable likelihoods. 
Table~\ref{tab:gen-model-compare} summarizes these distinctions across the model families.

This implicit–explicit distinction anticipates the foundational separation between 
\emph{evaluation} and \emph{generation} that we develop in Section~\ref{sec:probframe}. Explicit models 
naturally support evaluation through likelihoods or bounds, whereas implicit models are 
more closely aligned with the generation task itself. This connection becomes central to 
the probabilistic framework discussed next.

\section{Foundations of Generative AI}

The brief review of major generative modeling paradigms makes it clear that generative AI is a thriving area where researchers have made deep contributions to modeling and implementation. However, the foundations of generative AI have not been examined as deeply as one might hope. This might seem like a strange claim, given that many existing methodologies are firmly grounded in probabilistic modeling. But as we noted in our discussion of definitions in Section~\ref{sec:defs}, the task specification of generation does not appear to involve probability theory in any essential way.

In fact, there is a useful parallel here with the theory of prediction. While early formulations of prediction were couched in probabilistic terms, such as those of Vapnik and Valiant, later developments showed that game-theoretic and adversarial perspectives could also support rigorous foundational treatments. Today, we have several viable theories of prediction, including probabilistic and game-theoretic ones, and can choose among them depending on the setting. Could the theory of generation evolve in a similar way?

We think the answer is yes. Before outlining how theories of generation could be developed along both probabilistic and game-theoretic lines, let us first consider the benefits of adopting a foundational mindset toward generation.

\subsection{Why foundations?}

Reflecting on the foundations of an emerging field can lead to deeper insights into its core tasks and goals. While methods are central to implementation and practice, conceptually, the problem formulation must come first, followed by the development of methods that solve it. A clearer understanding of generation as a task may reveal where current methods fall short. Such recognition, in turn, can inspire the next generation of models.

A problem-first perspective also helps us avoid conflating generative AI with deep generative models. Even in settings where deep learning is dominant, such as prediction and reinforcement learning, we distinguish the task from the methods. No one would define prediction solely in terms of deep neural networks. Similarly, we should resist defining generative AI exclusively in terms of deep learning architectures. By abstracting away from implementation and focusing on underlying concepts, we hope to uncover deeper insights into both generative tasks and generative methods.

Beyond clarifying the task itself, a theoretical perspective on generation can also illuminate connections with other core problems in machine learning and statistics such as prediction, compression, and sequential decision making. We will return to these relationships in Section~\ref{sec:rel}.

Moreover, theory has often yielded unexpected benefits in related areas such as prediction. For instance, it is now known that online learning is provably harder than statistical learning: the Littlestone dimension can be strictly larger than the VC dimension, and yet we also have online-to-batch conversion theorems that allow us to convert online learners into batch learners with strong guarantees. Even more surprisingly, it has been shown that private learnability in binary classification is equivalent to online learnability. This result established a deep and unexpected connection between two seemingly unrelated fields: differential privacy and adversarial online learning. Finally, foundational work in online convex optimization led to the development of AdaGrad, which in turn inspired Adam, the optimizer implemented in PyTorch and TensorFlow and used across virtually all of deep learning today. These examples show that abstract theoretical frameworks can lead to both conceptual clarity and practical impact, a pattern we believe will carry over to generative modeling.

Finally, as Belkin~\cite{belkin2024necessity} has argued, we need theory to ensure that AI systems are safe, robust, controllable, and aligned with human values. While his argument concerns AI more broadly, it applies just as urgently to generative AI. These systems produce high-dimensional, open-ended outputs, and strong performance on benchmark datasets does not guarantee the absence of unexpected failure modes. A theoretical understanding is therefore necessary, though not sufficient, for the safe deployment of AI systems in human society.

\subsection{Probabilistic Framework}
\label{sec:probframe}

The probabilistic framework for studying generation is a classical one. We assume that there exists a probability distribution $p(\bx)$ that generates the training examples. Learning such distributions has long been a central topic in probability and statistics, though traditionally approached from a density estimation perspective. However, as we explain below, generating samples from a distribution and estimating its probability density are two distinct learning tasks. After clarifying this foundational distinction, we review theoretical results on the PAC learnability of distributions. We then turn to alternative, game-theoretic formulations of generation and to connections between generation, prediction, compression, and sequential decision making.

\subsubsection{Generation vs. Density Estimation}

Density estimation is a classical topic in statistics. It was formalized in the parametric setting through the foundational work of Fisher and Wald, and extended to nonparametric contexts by researchers such as Parzen and Rosenblatt. It also plays a central role in Vapnik's statistical learning theory, where it is listed as one of the three main learning problems,\footnote{The other two being ``pattern recognition" (prediction with binary functions) and ``regression estimation" (prediction with real-valued functions).} alongside classification and regression.

Formally, density estimation involves approximating a function \( p(\bx) \) that can be evaluated or integrated, while generation refers to producing new samples \( \bx \sim p(\bx) \) that resemble those drawn from the data distribution. Although related, these tasks are not equivalent. As we will see below, there are settings where one task is computationally easy while the other is provably hard under cryptographic assumptions.

This asymmetry appears in modern generative modeling as well. GANs define a distribution implicitly through a generator network and do not support density evaluation. VAEs maintain an approximate density, but only via a variational lower bound. Score-based models can produce high-quality samples using estimates of the score function \( \nabla_{\bx} \log p(\bx) \), but recovering the density itself is typically intractable. In contrast, autoregressive models and normalizing flows support both sampling and density evaluation, offering more direct access to the underlying distribution.

This distinction between generation and density estimation is not only conceptually important, but also practically significant. A model’s ability to sample does not imply it can reliably assign probabilities, and vice versa. For both theoretical analysis and real-world deployment, it is important to treat these as fundamentally different learning tasks. This distinction was formalized by Kearns et al.~\cite{kearns1994learnability} in their pioneering work on PAC learning of distributions, which we now review.

\subsubsection{PAC Learning of Distributions}

Valiant's original Probably Approximately Correct (PAC) learning framework was developed for binary classification. In a seminal paper, Kearns et al.~\cite{kearns1994learnability} extended this framework to study the learnability of probability distributions. Just as Valiant's PAC model introduced computational constraints to Vapnik's statistical learning theory~\cite{valiant1993view}, Kearns et al. revisited the classical literature on density estimation with a computational perspective.
One immediate consequence of this shift is the need to distinguish between 
\textbf{evaluation} and \textbf{generation}. 

In the \emph{evaluation model}, the learner outputs a probability mass function\footnote{For 
discrete domains. In continuous domains, a probability density function is appropriate.} 
$\hat p \in \cP$ that approximates the true distribution $p^\star$ in the sense that 
$\hat p(\bx)$ is close to $p^\star(\bx)$ for all $\bx$. 

In the \emph{generation model}, the learner uses random bits $\bb \in \{0,1\}^r$ to 
produce a generator $\hat g$ such that the distribution of $\hat g(\bb)$ approximates 
$p^\star$.

Kearns et al.\ quantified approximation using KL divergence. In the evaluation case, the goal is to ensure $KL(p^\star \,\|\, \hat p) \le \epsilon$. In the generation case, we require the same guarantee, but $\hat p$ now refers to the distribution induced by sampling $\hat g(\bb)$ when $\bb$ is drawn uniformly from $\{0,1\}^r$.

When computational resources are unrestricted, the two representations—an evaluator and a generator—can be converted into one another. Given $\hat p$, we can sample using techniques such as inverse transform sampling or rejection sampling. Conversely, given a generator $\hat g$, we can define the evaluator explicitly as
\[
\hat p(\bx) = \frac{|\{ \bb \in \{0,1\}^r : \hat g(\bb) = \bx \}|}{2^r} \ .
\]
However, these conversions are generally inefficient in high dimensions because the domain size grows exponentially with the number of bits.

A Boolean circuit with $n$ output bits defines a distribution over $\{0,1\}^n$ when its inputs are set to uniformly random bits. For the class of circuits consisting of $n$ OR-gates with small fan-in $k$, Kearns et al.\ showed that an efficient evaluator does not exist under a standard complexity-theoretic assumption,\footnote{Specifically, assuming that $\#P$ does not admit polynomial-size circuits.} even though the class is efficiently learnable with a generator for any constant value of $k$.

On the other hand, Naor~\cite{naor1996evaluation} constructed a class of distributions for which evaluation is efficient but generation is believed to be hard. In his construction, the probability of any given outcome can be computed efficiently using public information, but generating new valid samples requires inverting a cryptographic function, which is believed to be intractable without access to a secret key.

Framing distribution learning within the PAC paradigm naturally raises the question of whether there exists a complexity measure for distribution classes that can characterize sample complexity. In the standard PAC setting for classification, the VC dimension provides such a characterization. Diakonikolas~\cite{diakonikolas2016learning} posed this question explicitly in Open Problem 1.5.1:
\begin{quotation}
    Is there a “complexity measure” of a distribution class $\cP$ that characterizes the sample complexity of learning $\cP$?
\end{quotation}

Recent work by Lechner and Ben-David~\cite{lechner2024inherent} answers this question negatively under total variation distance. They show that no scale-invariant complexity measure, defined solely in terms of the distribution class, can characterize the sample complexity of PAC distribution learning. This holds even in the realizable setting. Their result applies to both the evaluation and generation models, as it concerns sample complexity alone and does not depend on computational considerations. This suggests that, unlike classification, distribution learning may not admit a single unifying notion of dimensionality that predicts sample efficiency in general. However, their result does not rule out the possibility of a scale-sensitive complexity measure that characterizes distribution learnability, nor does it address metrics other than total variation.

While the probabilistic framework focuses on learning a fixed data distribution, an alternative view frames generation itself as the outcome of a game between a learner and an adversary. We now turn to such game-theoretic formulations.

\subsection{Game-Theoretic Framework}

As we saw above, there is a well-developed approach to generation based on probabilistic foundations. But recall the definitions of generation from Section~\ref{sec:defs}. We observed that they shared some common features: use of training data, support for multiple modalities, similarity of generated data to training data, novelty of generated outputs, along with potential control over outputs via prompts.

It is interesting that none of these shared features make any explicit reference to probability theory. Perhaps we can have an alternative foundation for generation that is rooted in the interaction between one entity providing training examples and another entity using those training examples to generate new examples. Following the convention in learning theory, we will call the two entities the \emph{adversary} and the \emph{learner} respectively.

If interaction between the adversary and learner is going to be central to our modeling, it is no surprise that we will use game theory rather than probability theory as our foundation. There is precedent for doing so. For example, for the prediction problem there is a well-developed foundation based on game theory (see~\cite{cesa2006prediction} for an excellent introduction). In fact, probability theory itself can be built out of game-theoretic rather than measure-theoretic foundations~\cite{shafer2019game}.

The game-theoretic KM model we describe below is due to Kleinberg and Mullainathan~\cite{kleinberg2024language}. It is so natural that it almost falls out of the common features of definitions of generation. We will assume that there is an abstract space $\cX$ in which both training examples and generated examples will live. This space can be the space of strings, images, graphs, etc. The adversary chooses a set $S^\star \subseteq \cX$ ahead of the game without revealing it to the learner. The requirement of generating content similar to the training data will be formalized as the requirement of \emph{validity}: the learner should learn to generate objects in $S^\star$ eventually. The requirement of \emph{novelty} is simple to formalize: the generated object cannot be the same as anything the adversary has shown to the learner so far. We can collect all this into an online protocol that looks as follows.

\medskip

\noindent
For $t = 1, 2, \ldots$
\begin{itemize}
    \item Adversary plays $\bx_t \in S^\star$ 
    \item Learner plays $\hat \bx_t \notin \{ \bx_1, \ldots, \bx_t \}$
\end{itemize}

Note that there is no formal requirement for the learner to not repeat its own previous moves. But an adversary can always prevent that by first playing the learner's previous move and then following it up with its own move. This bare-bones protocol has surprisingly rich possibilities for theoretical analysis. But we do need to add a few more elements to complete the set up. First, the adversary can choose the order of elements adversarially but cannot withhold elements of $S^\star$ indefinitely. It must enumerate that full set. That is,
\[
\cup_t \{ \bx_t \} = S^\star \ .
\]
Second, following the learning theory tradition, we will assume that although the learner does not know which $S^\star$ the adversary picked at the start of the game, it knows the \emph{class $\cS$ of sets} from which $S^\star$ was picked. Third, we do not want the learner to run out of valid moves. So we will assume that every set $S$ in $\cS$ is infinite. Finally, we fix the following condition to determine whether the learner succeeds in generating novel yet valid objects:
\[
\exists t^\star < \infty \text{ such that } \forall t \geq t^\star,\ \hat\bx_t \in S^\star\ .
\]
In words, after some finite time point, the learner produces only valid objects.

The model described above was referred to as ``language generation in the limit" by Kleinberg and Mullainathan. The name comes from one of the earliest contributions to formal learning theory by Gold~\cite{gold1967language} who proposed a model called ``language identification in the limit". In Gold's model, the learner has to identify which $S^\star$ is generating the training examples. In contrast, in the KM model the learner simply has to output novel string from $S^\star$. Obviously, generation is easier than identification. Once the learner identifies the true $S^\star$ then, assuming it is infinite, it can output an unseen element of it. However, identification is known to be quite hard. In fact, Angluin~\cite{angluin1980inductive} characterized the classes that are identifiable in the limit. This characterization rules out most interesting classes. In contrast, Kleinberg and Mullainathan showed that \emph{any} countable class of sets is generatable in the limit! So generation is significantly easier than identification, an insight which is not possible without formal analysis of learning models.

The time $t^\star$ beyond which the learner produces valid outputs can depend on both the set $S^\star$ as well as the specific adversarially chosen enumeration of the set. We can strengthen this requirement by insisting that once $t^\star$ distinct elements of $S^\star$ have been enumerated, the learner must produce valid objects. This is called \emph{non-uniform generation}. In fact, it can be shown that all countable classes are generatable in this stronger non-uniform sense~\cite{li2025generation,charikar2025exploring}. Finally, \emph{uniform generation} means that the number $t^\star$ of distinct elements needed to be seen before valid generation cannot depend on $S^\star$ and must therefore only depend on the class $\cS$. Not all countable classes are uniformly generatable but there is a combinatorial dimension, called the \emph{closure dimension}, that characterizes uniform generatability~\cite{li2025generation}.

Another aspect of the KM model that has attracted attention is the trade-off between \emph{validity} and \emph{breadth}. Note that in the basic KM model, the learner simply has to generate novel valid objects. There is no requirement that the generated objects cover all or even a large part of $S^\star$. Researchers have therefore come up with various notions \emph{breadth} to formally study the trade-off between validity and breadth. This is not just an academic question. LLM hallucinations can be thought of as a violation of validity whereas mode collapse in GANs can be thought of as a lack of breadth. One motivation for studying validity-breadth trade-offs in a formal setting is to gain additional insights into practical problems such as hallucinations and mode collapse. For a broad overview of the KM model and subsequent developments including work on the validity-breadth tradeoff, we refer the reader to the COLT 2025 tutorial on Language Generation in the Limit~\cite{LangGenCOLT2025}.

\subsection{Post-Training of Generative Models}

Generative models are typically put to many downstream uses. In particular, one common use case is what Kevin Murphy calls ``generative design"~\cite[Chapter 20]{pml2Book}. Here, we use the generative model to find objects with desired properties. For example, a protein generator might be used to find enzymes that catalyze specific desired reactions. An LLM might be used to find proofs of a mathematical statement. 

In some cases, it might be possible to get to the desired answer just by using the right prompts. For example, adding ``let's think step by step" to a prompt can help LLMs solve reasoning tasks much better than a standard prompt~\cite{kojima2022large}. But for harder tasks, a generative models might need further training phases often referred to, especially in the LLM context, as post-training. In fact, LLMs are almost universally trained in multiple phases. After pretraining to maximize the likelihood of observed data, they typically undergo an instruction tuning phase which involves fine-tuning the base model with supervised data in the form of instruction, response pairs. A final stage is RLHF (reinforcement learning with human feedback) where human feedback on generated content is used to align the model to human values and preferences.

There is little theoretical understanding of this stage-wise training pipeline. More research is needed to understand the role each stage has in the final behavior of the model. We also need to better understand the sequence in which various behaviors are induced or reinforced. We have fairly clean abstraction, both probabilistic and game-theoretic, of the pretraining phase. It remains to be seen if similar abstractions exist for other training stages.

A lot of recent work on LLMs has gone into enhancing their reasoning capabilities especially in coding and math problems. An important post-training stage for developing reasoning LLMs is RLVR (reinforcement learning with verifiable rewards). This is a variation on RLHF where instead of human feedback, an LLM is fine-tuned to maximize a verifiable reward, i.e., an unambiguous signal that indicates correctness of the LLM's output. Examples of verifiable rewards include unit tests for code generation, numerical equality checks for mathematical questions, and other programmatically checkable criteria that unambiguously certify correctness. RLVR is often used in conjunction with CoT (Chain of Thought) traces where the LLM's output includes a full trace of the reasoning steps that led to the final answer. Sometimes there is a distinction made between outcome-based rewards, which only reward getting the final answer correct, and process-based rewards, which also reward partial progress towards the correct solution.

RLVR and CoT-based training has been the subject of some recent theoretical work that use learning-theory tools to shed light on reasoning LLMs.
Joshi et al.~\cite{joshi2025theory} study a model of autoregressive CoT generation in which a single, time-invariant ``step function" is applied repeatedly to produce a reasoning trace. They analyze the sample and computational complexity of learning this step function in settings where intermediate steps are observed or latent. One of their main results is that time-invariance provides sample complexity which is independent of the chain length. They also show that attention mechanisms emerge naturally in this framework.
Balcan et al.~\cite{balcan2025learning} address a complementary question: in RLVR pipelines, verifiers are often assumed to be given, but in practice they must themselves be learned from data. For example, if reasoning traces are in natural language then there may not be a formal checker. They formalize this problem in a PAC-learning framework, characterizing when such verifiers can be learned from labeled correct and incorrect traces, and proving both sample-complexity upper bounds and impossibility results under different data-access assumptions.

\subsection{Relationship with other machine learning areas}\label{sec:rel}

Prediction is intimately related to generation. Autoregressive models reduce generation to a sequence of next token prediction problems. GANs use a classifier to improve the generator. Diffusion models use a sequence of de-noising predictors that learn to remove noise given slightly noisy versions as input. More work is needed to understand the nature of these connections. For example, one of the algorithms of Kearns et al.~\cite{kearns1994learnability} for learning parity gate distributions uses a PAC learning algorithm for parity functions as a subroutine. Developing general reduction theorems that transform a prediction algorithm with known guarantees into a generation algorithm with corresponding generative guarantees remains an open challenge. While the focus here is on using prediction to enable generation, the reverse direction, using generation to aid prediction, also appears in many settings such as data augmentation and missing data imputation.

In prediction, models with succinct description generalize well to unseen data~\cite{blumer1987occam}. Conversely, for any PAC learnable class, we can describe an $\epsilon$-accurate hypothesis using only $O(\log(1/\epsilon))$ bits. Exploring how compression and generation are related is a fascinating open area. Following up on a conjecture of Kearns et al~\cite{kearns1994learnability}, Naor~\cite{naor1996evaluation} showed that, in sharp contrast to PAC learning, there is a class of efficiently generatable distributions where any $\epsilon$-accurate efficient generator needs to be of size $\Omega(1/\epsilon)$. However, the construction uses pseudo-random generators leaving open the possibility that succinct generators may exist for natural distribution families. Many other analogs of the rich relationship between prediction and compression remain to be explored in the generative setting, including the role of sample compression schemes~\cite{floyd1995sample,moran2016sample}.

Reinforcement learning (RL) is already an integral component in the training pipelines of generative models, particularly large language models (LLMs). Approaches such as Reinforcement Learning with Human Feedback (RLHF) and Reinforcement Learning with Verifier Feedback (RLVR) fine tune base models to align them with human preferences and to enhance their reasoning capabilities. Conversely, generative AI can also significantly advance reinforcement learning. Offline RL requires data while online RL requires an environment. Generative models can provide synthetic data~\cite{wang2024pre} for offline RL and simulated environments, or world models~\cite{ha2018world,matsuo2022deep}, to facilitate online RL. In fact recent work suggests that LLMs may in fact already be learning world models implicitly~\cite{vafa2024evaluating}.

More fundamentally, reinforcement learning itself can be viewed as generative modeling focused explicitly on producing behavior sequences that achieve high rewards. Generating behaviors conditioned on achieving high rewards is thus a natural and promising approach to RL. Recent methods such as Decision Transformer~\cite{chen2021decision} and Decision Diffuser~\cite{ajay2023conditional} explicitly adopt this viewpoint. However, conditioning generation exclusively on high rewards raises foundational challenges related to trajectory realizability, particularly in stochastic environments where agents cannot directly control states, only actions. Addressing fundamental theoretical questions, such as characterizing limits of this generative and RL interaction, remains an open and fertile area for future research. An emerging line of work at this intersection is Generative Flow Networks, which we discuss next.

An emerging example that explicitly merges ideas from generative modeling and reinforcement learning is the family of Generative Flow Networks (GFlowNets)~\cite{bengio2023gflownet}. A GFlowNet learns a stochastic policy for constructing objects step by step, with the objective of sampling them with probability proportional to a given reward function. This framing treats generation as a reinforcement learning problem over compositional action spaces, while also providing a probabilistic model over the final outcomes. Unlike argmax-based optimization, GFlowNets sample outcomes proportional to their rewards, enabling the discovery of diverse high-reward solutions rather than a single mode.

\subsection{Socially Responsible Generation}

AI, especially generative AI, is advancing rapidly. Generative AI technology is now widely deployed, and its societal impact is expected to grow. It is therefore imperative that generative techniques be developed with respect for human rights and for the broad benefit of society, not just a narrow segment. For example, humans have a right to keep their personal information private and to know if they are interacting with an AI agent or working with AI-generated content. Creators have a right to own the content or intellectual property they produce. Many of these issues are unsettled, and progress will require adjustments to existing political, legal, and societal structures. These problems are not purely technical. Still, a theoretical framework can help clarify what is technologically and scientifically feasible. We will discuss some technical questions related to privacy and to watermarking and detection of AI-generated content. We will briefly touch upon copyright and IP, which lie largely outside the scope of this chapter.

\subsubsection{Privacy}

Differential privacy (DP) is a well‐established theoretical framework within which machine learning algorithms with formal privacy guarantees have been developed and deployed. In DP learning there is a fundamental trade‐off between the strength of the privacy guarantee and the sample complexity needed to achieve a target accuracy: smaller privacy parameters (more stringent privacy) typically require more samples, and in some cases can render an otherwise learnable problem unlearnable. For example, threshold functions are easy to learn in the PAC framework without privacy, but cannot be learned with DP. There is a remarkable structural connection here with online learning: the classes that are learnable with approximate DP are exactly those that are online learnable. Approximate DP allows a small failure probability $\delta$, in contrast to pure DP, and this relaxation is essential for the equivalence with online learnability.

For distribution learning under the total variation metric, Bun et al.~\cite{bun24not} construct a class that is learnable without privacy but not learnable with approximate DP. However, they do not consider other metrics, nor do they investigate whether natural distribution families satisfy conditions under which private distribution learning is possible. There is much to be done in understanding the possibilities and limits of privately learning distributions for both evaluation and generation.

\subsubsection{AI-Generated Content Detection}

AI-generated content detection, especially AI-generated text detection~\cite{fraser2025detecting}, is an emerging area with broad and important applications. Related work on detecting AI-generated images and video, most prominently deepfakes, has explored similar statistical and machine learning approaches for distinguishing synthetic from natural content~\cite{mirsky2021creation}. Such systems can help enforce compliance with policies, for example by preventing the use of AI tools for reviewing academic papers or completing course homework. They can also support efforts to combat online misinformation by clearly identifying and tagging AI-generated material. Finally, detection tools can benefit AI development itself, since training AI models on AI-generated content without safeguards can lead to model collapse. In what follows, we focus on the theoretical and methodological foundations of AI content detection.

AI content detection strategies can be broadly divided into two categories: those that require cooperation from builders of AI systems and those that do not. The first category includes approaches such as watermarking, where a cryptographic signature is embedded in the generated output and can later be used to verify provenance. The second category includes approaches that use machine learned classifiers or statistical tests to distinguish naturally occurring objects from AI-generated ones. These categories have distinct strengths and weaknesses: cooperative watermarking can be highly reliable under the right conditions but is less flexible, whereas post-hoc watermark-free detection is more adaptable but tends to be less reliable. 

Let us adopt a statistical hypothesis testing viewpoint to understand both watermark-free and watermark-based approaches from a unified perspective. Suppose $X$ is a digital object whose provenance we wish to determine. In the watermark-free case (see, e.g.,~\cite{radvand2025zero}), we test
\[
H_0: X \sim \pnat, \quad H_1: X \sim \pai,
\]
where $\pnat$ is the distribution of non-AI-generated content (e.g., natural images, human-written text) and $\pai$ is the distribution of AI-generated content. This matches the compliance setting in which $H_0$ represents compliance and $H_1$ represents a violation. We aim to keep the false positive (Type I) rate below a small threshold (say $1\%$) while minimizing the false negative (Type II) rate. The main challenge for watermark-free detection is that AI systems are increasingly good at mimicking $\pnat$. Nothing prevents $\pai$ from approaching $\pnat$ arbitrarily closely, creating a cat-and-mouse dynamic: as generators improve, the separation between $\pai$ and $\pnat$ shrinks, and tests that rely on a fixed degree of separation lose reliability. In the watermarking case, by contrast, the generator’s distribution is intentionally altered to create a detectable signal, allowing stronger and more stable tests.

In the watermarking case (see, e.g.,~\cite{li2025statistical}), the generator is modified to produce outputs from a distribution $\paik$ that embeds a statistical signal determined by a secret key $k$. This key is used to generate a pseudorandom sequence that guides the generation process, for example by biasing the selection of certain output elements. The resulting hypothesis test is
\[
H_0: X \sim \pnat, \quad H_1: X \sim \paik,
\]
where $\paik$ is constructed to be close to $\pai$ in terms of utility or perceptual quality, yet to differ in a controlled way that is detectable given access to $k$. With knowledge of $k$, one can compute a detection statistic whose null distribution is known or can be reliably estimated and whose distribution under $H_1$ is shifted to enable reliable detection. This intentional separation between $\pnat$ and $\paik$ allows for tests with provable Type~I and Type~II error guarantees, and enables detection performance to remain stable even as generative models improve.

We have focused most of our discussion above on the detection side. In practice, however, adversarial actors may attempt to evade detection, especially if details of the detection mechanisms are public. For watermarking, an adversary might seek to remove the watermark by applying transformations to the watermarked text. For watermark-free detection, an adversary might fine-tune the AI model to minimize a known detection score or add perturbations to the generated object to fool the test. From a theoretical viewpoint, this motivates questions about optimal strategies in a sequential game in which the detector designs tests and the adversary adapts to evade them. Addressing such questions will be critical for designing future detection systems that remain robust in adversarial settings.

\subsubsection{Copyright and IP}

As a transformative technology, generative AI also raises important questions about copyright and intellectual property (IP). These arise on both the input and output sides of generative models. On the input side, there are ongoing lawsuits concerning the use of copyrighted material in the training sets of generative models. On the output side, individuals have filed copyright claims for AI-generated works created in response to human prompts. The regulatory and legal issues in these areas are far from settled. Because these issues are primarily legal rather than technical, we do not address them in detail here. We instead point the reader to recent \emph{Communications of the ACM} articles by Pamela Samuelson, which provide accessible and up-to-date overviews of the legal landscape~\cite{samuelson2025authorship,samuelson2024remedies,samuelson2024questions,samuelson2023challengesI,samuelson2023challengesII}.

\bibliographystyle{plain}
\bibliography{mybibfile}

\end{document}

%% file: macros.tex
\newcommand\bb{\mathbf{b}}
\newcommand\bc{\mathbf{c}}
\newcommand\bu{\mathbf{u}}
\newcommand\bx{\mathbf{x}}
\newcommand\bz{\mathbf{z}}
\newcommand\bepsilon{\bm{\epsilon}}

\newcommand\cN{\mathcal{N}}
\newcommand\cP{\mathcal{P}}
\newcommand\cS{\mathcal{S}}
\newcommand\cX{\mathcal{X}}

\newcommand\pnat{P_{\text{natural}}}
\newcommand\pai{P_{\text{AI}}}
\newcommand\paik{P_{\text{AI}}^{(k)}}

\newcommand\eye{\mathbf{I}}
\newcommand{\KL}{\mathrm{KL}}
\newcommand{\EE}{\mathbb{E}}
\newcommand\stdnormal{\mathcal{N}\!\big( \mathbf{0},\,\mathbf{I}\big)}

%% file: abstract.tex
Beginning with text and images, generative AI has expanded to audio, video, computer code, and molecules. Yet, if generative AI is the answer, what is the question? We explore the foundations of generation as a distinct machine learning task with connections to prediction, compression, and decision-making. We survey five major generative model families: autoregressive models, variational autoencoders, normalizing flows, generative adversarial networks, and diffusion models. We then introduce a probabilistic framework that emphasizes the distinction between density estimation and generation. We review a game-theoretic framework with a two-player adversary-learner setup to study generation. We discuss post-training modifications that prepare generative models for deployment. We end by highlighting some important topics in socially responsible generation such as privacy, detection of AI-generated content, and copyright and IP. We adopt a task-first framing of generation, focusing on what generation is as a machine learning problem, rather than only on how models implement it.